\documentclass[10pt, conference]{IEEEtran}
\usepackage[english]{babel}

\usepackage[dvipsnames,usenames]{xcolor}

\usepackage{colortbl}
\usepackage{comment}
\usepackage{graphicx}
\usepackage{epsfig}
\usepackage{array}           
\usepackage{listings}
\usepackage{epstopdf}
\usepackage{multirow}
\usepackage{subcaption}
\usepackage{rotating}
\usepackage{float}
\usepackage[obeyspaces,hyphens,spaces]{url}
\usepackage{balance}
\usepackage{fancybox}
\usepackage{scalefnt}
\usepackage[normalem]{ulem}  
\usepackage{hyperref}

\usepackage{pgfplots}
\pgfplotsset{compat=1.18}
\usepackage{booktabs}        
\usepackage{amssymb}         
\usepackage{pifont}          
\usepackage{amsmath}
\usepackage{tcolorbox}
\usepackage{algorithm}
\usepackage{algpseudocode}

\usepackage[
  disable,
  colorinlistoftodos
]{todonotes}

\usepackage{verbatim}


\title{Towards Assessing Deep Learning Test Input Generators}
\author{
  \IEEEauthorblockN{Seif Mzoughi\IEEEauthorrefmark{1}, Ahmed Haj yahmed\IEEEauthorrefmark{1}, Mohamed ELSHAFEI\IEEEauthorrefmark{1}, Foutse khomh\IEEEauthorrefmark{1}, Diego Elias Costa \IEEEauthorrefmark{2}}
  \IEEEauthorblockA{\IEEEauthorrefmark{1}Polytechnique Montreal\\
  Email: \{Seif.mzoughi, ahmed.haj-yahmed, mohamed.elshafei, foutse.khomh\}@polymtl.ca}
  \IEEEauthorblockA{\IEEEauthorrefmark{2}Concordia University\\
  Email: diego.costa@concordia.ca}
}
\begin{document}
\maketitle
\begin{abstract}

Deep Learning (DL) systems are increasingly deployed in safety-critical applications, yet they remain vulnerable to robustness issues that can lead to significant failures. While numerous Test Input Generators (TIGs) have been developed to evaluate DL robustness, a comprehensive assessment of their effectiveness across different dimensions is still lacking. This paper presents a comprehensive assessment of four state-of-the-art TIGs—DeepHunter, DeepFault, AdvGAN, and SinVAD—across multiple critical aspects: fault-revealing capability, naturalness, diversity, and efficiency. Our empirical study leverages three pre-trained models (LeNet-5, VGG16, and EfficientNetB3) on datasets of varying complexity (MNIST, CIFAR-10, and ImageNet-1K) to evaluate TIG performance. Our findings reveal important trade-offs in robustness revealing capability, variation in test case generation, and computational efficiency across TIGs. The results also show that TIG performance varies significantly with dataset complexity, as tools that perform well on simpler datasets may struggle with more complex ones. In contrast, others maintain steadier performance or better scalability. This paper offers practical guidance for selecting appropriate TIGs aligned with specific objectives and dataset characteristics. Nonetheless, more work is needed to address TIG limitations and advance TIGs for real-world, safety-critical systems.

\end{abstract}

\section{Introduction}
\label{sec:introduction}
Deep Learning (DL) has emerged as a transformative solution across application domains, extracting complex patterns from diverse data sources~\cite{he2016deep, young2018recent, geoffrey2012deep}.
DL components are now integral to safety-critical systems such as self-driving cars~\cite{bojarski2016end} and aircraft collision avoidance systems~\cite{julian2016policy}, where reliability and robustness are paramount.
 

However, despite their growing adoption, DL systems struggle to maintain consistent performance under diverse conditions. A key concern is robustness - the system's ability to maintain reliable performance when faced with variations in input data that differ from training conditions. For instance, an autopilot system's fatal accident~\cite{noauthor_ubers_2020} occurred due to failure in handling lighting conditions different from its training data. Other DL system failures stem from different issues: a facial recognition system's misidentification~\cite{sarlin_false_2021} highlighted accuracy limitations, while hiring system bias against women~\cite{dastin_amazon_2018} revealed fairness concerns. These incidents collectively demonstrate that high accuracy on standard datasets does not guarantee reliable real-world performance, with robustness being a particularly critical concern for safety-critical applications.
To address these robustness challenges, recent efforts have focused on developing test input generators (TIGs)\cite{tian2018deeptest, pei2017deepxplore, xie2019deephunter}. TIGs are specialized techniques designed to generate new test inputs—either by modifying existing data or creating novel instances—that rigorously evaluate and stress-test DL models. They expose vulnerabilities unique to DL systems, such as sensitivity to minor input perturbations and susceptibility to adversarial examples, which traditional test case generation methods fail to uncover. TIGs specifically target these weaknesses by crafting inputs that can cause the model to produce erroneous outputs, thereby revealing robustness issues. 

Over the years, numerous TIGs have been developed, leveraging a wide range of strategies including (1) pixel-level perturbation \cite{shah2020efficient}; (2) manipulation of the input representation using generative DL models such as generative adversarial networks (GANs)  \cite{goodfellow2020generative} or variational autoencoder (VAE) \cite{pinheiro2021variational}. These TIGs vary widely in their methodologies and focus areas. Some focus on applying subtle perturbations to existing data, others generate entirely new data instances, and each introduces unique techniques to challenge and evaluate the robustness of DL models in different ways.

Despite the prevalence of TIGs, several challenges persist in the field. Researchers and developers now have access to a broad spectrum of tools, each operating differently, yet there has been no comprehensive assessment of TIGs across various models and datasets. This lack of evaluation hinders the generalization of their effectiveness and makes it challenging for practitioners to select the most suitable tool for their specific needs. The community requires deeper insights into these TIGs to better understand their capabilities and limitations.

Furthermore, clear assessment criteria for evaluating TIGs remain undefined. Many studies focus on generating failure-inducing cases without adequately considering the quality of the generated data. Assessing this quality, especially in terms of naturalness \cite{li2023towards}, is a significant challenge. Simple rule-based or distance-based metrics often fail to capture the nuances of human perception, leading to test cases that may not reflect real-world scenarios and thus limiting their practical utility. Additionally, efficiency in test case generation is often neglected, even though it is crucial for real-world applicability. Many TIGs are computationally intensive and resource-demanding, which hampers their use in practical settings. These challenges highlight the need for more comprehensive evaluation criteria that consider not only the fault-revealing capabilities of TIGs but also the naturalness, diversity, and efficiency of the generated test cases.


In this paper, we undertake the first comprehensive assessment of TIGs for DL systems, aiming to identify and understand their effectiveness across multiple dimensions: fault-revealing capability, naturalness, diversity, and efficiency. This paper guides selecting appropriate TIGs based on specific testing needs and resource constraints while establishing a framework for assessing and improving testing approaches. To guide our investigation, we formulate our research questions as follows:

\begin{itemize}
    \item \textbf{RQ1:} Which TIG \textbf{reveals} more DL robustness issues?

    \item \textbf{RQ2}: Which TIG generates more \textbf{natural test cases}?
    
    \item \textbf{RQ3}: Which TIG generates more \textbf{diversified test cases}?

    \item \textbf{RQ4:} Which TIG is more \textbf{efficient} in test case generation?

\end{itemize}

We conduct an empirical study evaluating four state-of-the-art TIGs: DeepHunter \cite{xie2019deephunter}, DeepFault \cite{eniser2019deepfault}, AdvGAN \cite{xiao2018generating}, and SinVAD \cite{kang2020sinvad}. To ensure the generalizability of our findings, we leverage three DL models of varying complexities (LeNet-5 \cite{lecun1998gradient}, VGG16 \cite{simonyan2014very}, and EfficientNetB3 \cite{shah2020efficient}) applied to three datasets of different sizes (MNIST \cite{lecun1998gradient}, CIFAR-10 \cite{_krizhevsky_cifar-10_nodate}, and ImageNet-1K \cite{deng2009imagenet}). We assess the TIGs on four dimensions: fault-revealing capability, naturalness \cite{li2023towards}, diversity, and efficiency. These dimensions represent the critical aspects that determine a TIG's practical value \cite{zhang2020machine}. In fact, fault-revealing capability demonstrates effectiveness in identifying real issues \cite{pei2017deepxplore}, naturalness ensures that test cases reflect realistic scenarios \cite{li2023towards}, diversity guarantees comprehensive testing coverage \cite{ma2018deepgauge}, and efficiency enables practical deployment. Based on this evaluation framework, we aim to provide practical insights for both the software engineering (SE) and artificial intelligence (AI) communities on the selection and application of TIGs for robustness testing.

To summarize, this paper makes the following contributions:
 \begin{itemize}
  \item We present a comprehensive empirical study assessing four state-of-the-art TIGs—DeepHunter, DeepFault, AdvGAN, and SinVAD—across four dimensions, including fault-revealing capability, naturalness, diversity, and efficiency.
    \item Our results (1) highlight trade-offs among the TIGs in robustness-revealing capability, test case variation, and efficiency, and (2) reveal significant performance shifts based on dataset complexity, with each tool showing varying behavior across different evaluation scenarios.
  
  
  \item We offer actionable recommendations for practitioners and researchers for selecting appropriate testing tools based on specific testing objectives and dataset characteristics.

  \item We release a replication package \cite{replicationpackage}, including our detailed results, that can be used as a benchmark for other studies on DL robustness testing.
  
\end{itemize}


The remainder of this paper is structured as follows: Section II outlines the key concepts employed throughout our work. Section III describes the study design. Section IV summarizes the evaluation findings across research questions RQ1 to RQ4. Section V presents a discussion of the results, and Section VI explores related work. Section VII examines threats to the study's validity. Finally, Section VIII concludes the paper.

\section{Background}
\label{sec:background}
\subsection{Deep Learning Systems}

In simple terms, Deep Neural Networks (DNN) are neural networks with multiple layers of interconnected neurons~\cite{lecun2015deep}.
Their multiple layers of neurons allow DNNs to learn increasingly abstract representations of the input data, making remarkable achievements in a variety of applications, including image recognition~\cite{he2016deep}, natural language processing~\cite{young2018recent}, and speech recognition~\cite{geoffrey2012deep}.
A DL system is any software system with at least one DNN component~\cite{ma2018deepgauge}. 
DL systems vary from being designed purely through arrays of DNN components (e.g., object detection systems) or a combination of DNN components and traditional software components (e.g., recommendation systems).

Unlike a typical software system, where system behavior is governed by a previously defined business logic, a DL system derives its behavior from the input data. 
This data-driven nature significantly impacts the robustness of DL systems, making them sensitive to poor data quality and vulnerable to adversarial attacks, i.e., when a user deliberately manipulates input data to cause the system to exhibit erroneous behavior.

\subsection{Deep Learning Testing} 

Testing DL systems is an essential quality assurance task to ensure safe deployment, particularly in safety-critical domains. 
DL Testing is a set of practices that aims to discover erroneous behaviors in DL systems~\cite{zhang2020machine}, i.e., errors or unexpected behaviors caused by programming mistakes, model design flaws, or data quality and preprocessing concerns~\cite{zhang2020machine}. Due to their opaque black-box nature, testing large-scale DL systems with millions of neurons and thousands of parameters for all potential inputs is very difficult.

DL testing approaches are designed to assess various DL components and find different types of erroneous behaviors to meet a testing objective (like robustness). Robustness \cite{braiek2024machinelearningrobustnessprimer}
is the DL system's ability to correctly operate under stressful conditions~\cite{zhang2020machine}, such as varying changes in light exposure and angles of an image classification system. In the presence of disturbances, robust systems should sustain performance. However, robustness remains a critical challenge for DL systems. For example, in one tragic Tesla crash \cite{yadron_tesla_2016}, a white truck was missed due to the brightness of the sky.

\subsection{Test Input Generators} 
\label{sec:back_TIGs}

\begin{table*}[!t]
    \small
    \setlength{\tabcolsep}{7pt}
    \renewcommand{\arraystretch}{1.3}
    \caption{Comparison of Test Input Generators (TIGs)}
    \label{tab:TIG_characteristics} 
    \setlength{\tabcolsep}{3pt}  
\begin{tabular}
{>{\raggedright\arraybackslash}p{2cm}>{\raggedright\arraybackslash}p{1.5cm}>{\raggedright\arraybackslash}p{1.2cm}>{\raggedright\arraybackslash}p{2cm}>{\raggedright\arraybackslash}p{3cm}>{\raggedright\arraybackslash}p{2.8cm}>{\raggedright\arraybackslash}p{3cm}}
\toprule
\textbf{Category} & \textbf{Tool} & \textbf{Access Type} & \textbf{Key Objective} & \textbf{Generation Strategy} & \textbf{Data Quality \& Preservation} & \textbf{Notable Features} \\ 
\midrule
PBA (Perturbation-Based Approach) 
& \cellcolor{gray!10}DeepHunter \cite{wei2021deephunter} 
& \cellcolor{gray!10}White-box 
& \cellcolor{gray!10}Coverage Optimization 
& \cellcolor{gray!10}Coverage-guided fuzzing with metamorphic mutations 
& \cellcolor{gray!10}Rule-based, $L_0$ and L\textsubscript{$\infty$} constraints 
& \cellcolor{gray!10}Coverage Criteria: BKNC,KMNC,NBC,
NC, SNAC, TKNC metrics \\

& DeepFault \cite{eniser2019deepfault} 
& White-box 
& Fault Localization 
& Suspiciousness metrics: Tarantula, Ochiai, DStar 
& L\textsubscript{$\infty$} constraints 
& Targets suspicious neurons for fault localization \\
\midrule
GMA (Generative Model Approach)
& \cellcolor{gray!10}SinVAD \cite{kang2020sinvad} 
& \cellcolor{gray!10}Black-box 
& \cellcolor{gray!10}Diversity in Test Cases 
& \cellcolor{gray!10}VAE with Neuron Coverage 
& \cellcolor{gray!10}Domain-constrained inputs 
& \cellcolor{gray!10}Realistic, diverse inputs enhancing neuron activation \\

& AdvGAN \cite{xiao2018generating} 
& Black-box 
& Misbehavior 
& GAN 
& Adversarial loss 
& Single forward pass generation \\
\bottomrule
\end{tabular}

    \vspace{1mm}
\end{table*}

Test Input Generators (TIGs) play a crucial role in assessing the robustness and reliability of DL systems by generating 
failure-inducing examples. These artificial inputs, where the DL system's predicted label deviates from the expected one, expose vulnerabilities and misbehaviors in the system \cite{zohdinasab2024exposing}. To tackle the oracle problem, i.e., determining the expected label for a new input, TIGs often apply minor alterations to inputs with known labels, under the assumption that such perturbations do not alter the label. When misbehaviors are identified, the generated inputs can assist developers in refining the system.

Additionally, TIGs may address other test objectives \cite{huang2020survey} when exposing misbehaviors. Some TIGs prioritize coverage optimization, where the goal is to activate as many neurons \cite{aggarwal2018neural}, pathways, or architectural components \cite{khamparia2019systematic} as possible, maximizing the chance of exposing edge cases or latent issues. Others focus on input diversity \cite{pei2017deepxplore} as a key objective, as it encourages the generation of varied test cases that cover a broader spectrum of operational scenarios, thereby enhancing the robustness of the model.

TIGs may operate with white-box access to the DL system, allowing them to exploit information such as neuron activations and layer-specific responses to guide the generation of test cases. In contrast, with black-box access, TIGs treat the model as an opaque system and rely solely on input-output observations. While generally less configurable than white-box methods, black-box approaches are more adaptable and applicable across diverse model architectures.
We discuss two primary approaches for handling input images: Perturbation-Based and Generative Models \cite{10633581,riccio2023and}, highlighting representative tools from the literature and their characteristics as shown in Table \ref{tab:TIG_characteristics}.

\textbf{Perturbation-Based Approach (PBA)}: PBA approaches \cite{ivanovs2021perturbation} generate new inputs by applying small modifications, or “perturbations”, to existing seed inputs. TIGs in this category directly modify images in the original pixel space through perturbations of pixel values. These modifications assume the original label remains correct and serve to identify situations where minimal imperceptible changes can destabilize the model’s output.

For instance, DeepHunter \cite{xie2019deephunter} is a PBA approach leveraging coverage-guided fuzzing to detect defects in DL systems. By employing a metamorphic mutation \cite{ding2017validating} strategy and advanced seed selection techniques, DeepHunter generates new test cases and maximizes defect detection coverage \cite{sun2019structural}.DeepHunter uses metrics like Neuron Coverage (NC), K-Multisection Neuron Coverage (KMNC), Neuron Boundary Coverage (NBC), Strong Neuron Activation Coverage (SNAC), and Top-K Neuron Coverage (TKNC) to guide diverse test case generation and enhance defect detection. Using these coverage techniques, DeepHunter \cite{wei2021deephunter} introduces data perturbations such as pixel-level adjustments, brightness changes, and geometric transformations, ensuring that new tests remain valid while revealing hidden model vulnerabilities.

DeepFault \cite{eniser2019deepfault}, on the other hand, is a white-box fault-localization approach \cite{zheng2016fault} that identifies suspicious neurons contributing to model misbehavior. DeepFault uses fault-localization techniques like Tarantula \cite{jones2005empirical}, Ochiai \cite{ochiai1957zoogeographic}, and DStar \cite{wong2013dstar} to identify suspicious neurons contributing to model misbehavior. Each method assigns suspicion scores based on neuron activations in passing versus failing cases, with higher scores marking neurons more likely linked to faults, aiding precise debugging. It generates test inputs by applying targeted perturbations to maximize the activation of these identified neurons, helping to expose model weaknesses while providing insights into the network's decision-making process.

\textbf{Generative Model Approach (GMA)}: GMA approaches leverage techniques such as Generative Adversarial Networks (GANs) \cite{heusel2017gans} or Variational Autoencoders (VAEs) \cite{pinheiro2021variational} to synthesize entirely new inputs that mimic the distribution of the training data. 
By generating realistic but novel inputs, these methods allow for a broader evaluation of the DL model’s robustness, often uncovering unexpected behavior in regions of the input space that may not be reachable through perturbations alone.

Sinvad \cite{kang2020sinvad} is a GMA approach that combines VAEs with targeted search strategies to generate semantically valid test inputs. By learning the underlying distribution of valid inputs, Sinvad generates diverse test cases while maintaining the domain's semantic constraints. The approach employs a novel validation mechanism to ensure generated inputs remain realistic and meaningful while simultaneously maximizing neuron coverage and exploring diverse behavioral patterns of the DL system.

Another approach, AdvGAN \cite{xiao2018generating}, utilizes GANs to generate adversarial examples. It trains a generator network to create perturbations that, when added to legitimate inputs, cause misclassification while maintaining high perceptual quality \cite{choi2020deep}. Unlike traditional perturbation methods, AdvGAN learns to generate adversarial examples in a single forward pass, making it computationally efficient. The generator is trained using a combination of adversarial loss for realism, a fooling objective to induce misclassification, and a perceptual constraint to preserve visual similarity to the original inputs. 

Unlike PBA techniques, GMA approaches can produce realistic inputs that may differ significantly from the original dataset, as they operate in a compressed input space. However, the quality of generated inputs heavily depends on the training data and the effectiveness of the generative model. These methods offer a broader exploration of the input space, potentially uncovering misbehaviors that PBA techniques might miss due to their reliance on perturbations of existing inputs.

\section{Study Design}
\label{sec:approach}

\begin{figure*}[ht]
    \centering
    \includegraphics[width=\textwidth]{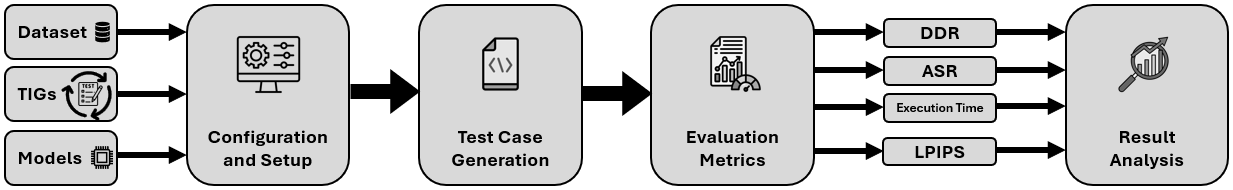}
    \caption{Study Methodology for Assessing Test Input Generator (TIG) Performance}
    \label{fig:workflow}
\end{figure*}

In this study, we evaluated DL TIGs using a three-step methodology: (1) selection of representative pre-trained models, datasets, and TIGs; (2)  use of TIGs to generate synthetic test inputs; and (3) evaluation of the TIGs. Figure \ref{fig:workflow} illustrates this workflow. First, we selected four well-known TIGs that represent the main categories in the literature: DeepHunter \cite{wei2021deephunter}, DeepFault \cite{eniser2019deepfault}, SinVad \cite{kang2020sinvad}, and AdvGAN \cite{xiao2018generating}. For DL models, following previous comprehensive evaluation studies of DL testing approaches that assess multiple TIGs across varied model architectures and datasets 
\cite{riccio2023and}, we chose three popular pre-trained models with varying levels of complexity: LeNet-5 \cite{lecun1998gradient} (small complexity), VGG-16 \cite{simonyan2015deepconvolutionalnetworkslargescale} (medium complexity), and EfficientNetB3 \cite{tan2020efficientnetrethinkingmodelscaling} (large complexity). To evaluate these models using TIGs, we employed three widely used publicly available datasets of different sizes: MNIST \cite{726791} (small size), CIFAR-10 \cite{_krizhevsky_cifar-10_nodate} (medium size), and ImageNet1K \cite{5206848} (large size).

We then applied each TIG to each pre-trained DL model to generate new input samples to test model robustness. Finally, we assessed each TIG using four key metrics: Defect Detection Rate (DDR) \cite{xie2019deephunter}, Attack Success Rate (ASR), Learned Perceptual Image Patch Similarity (LPIPS) \cite{zhang2018unreasonable}, and perturbation magnitude (PM)\cite{luo2018towards}. 
Each experiment was repeated ten times, and all values were recorded. We used averages for summarizing results and the raw values to conduct statistical significance tests on the differences between means.

The following sections describe the datasets selected, TIGs used for comparison, the DL models employed, and the metrics used in our experiments.

\subsection{Models and Datasets}
\label{sec:models_and_datasets}

We selected three widely recognized model-dataset pairs to assess TIGs, representing different complexity scales in DL. We chose LeNet-5 and VGG16, foundational architectures designed for image classification tasks. LeNet-5 processes $28 \times 28$ grayscale images and is paired with the MNIST dataset, which contains 60,000 training and 10,000 test images of handwritten digits across 10 classes. VGG16, a deeper Convolutional Neural Network (CNN) with 16 layers, handles $32 \times 32 \times 3$ RGB images and is suited for the CIFAR-10 dataset, comprising 50,000 training and 10,000 test images across 10 classes. To incorporate models with higher complexity, we included EfficientNetB3 from the family of efficient architectures \cite{shah2020efficient} using compound scaling \cite{xie2020adversarial}. EfficientNetB3 processes bigger resolution images of size $224 \times 224 \times 3$ and is paired with the ImageNet1K dataset. ImageNet1K offers a collection of 1.2 million training images and 50,000 validation images across 1,000 classes, making it a standard benchmark for image classification tasks. These dataset-model pairs represent evaluation subjects with increasing complexity and resolution: MNIST ($28 \times 28$ grayscale), CIFAR-10 ($32 \times 32$ RGB), and ImageNet-1k ($300 \times 300$ RGB), as detailed in Table \ref{tab:datasetso}. This progression allows us to evaluate the robustness of the TIGs across diverse data distributions and feature richness.

These combinations were selected based on three criteria: (1) established performance benchmarks in the literature, (2) varying architectural complexity to test TIG scalability, and (3) diverse image characteristics to evaluate TIG adaptability. All models used are pre-trained versions that achieve competitive accuracy on their respective datasets.

\begin{table*}[!t]
    \small
    \setlength{\tabcolsep}{7pt}
    \renewcommand{\arraystretch}{1.3}
    \caption{Comparison of Test Input Generators (TIGs)}
    \label{tab:datasetso}
    \begin{tabular}{>{\raggedright\arraybackslash}p{2cm}>{\raggedright\arraybackslash}p{2cm}>{\raggedright\arraybackslash}p{2.5cm}>{\raggedright\arraybackslash}p{2cm}>{\raggedright\arraybackslash}p{2cm}>{\raggedright\arraybackslash}p{2cm}>{\raggedright\arraybackslash}p{2cm}}
\toprule
\textbf{Dataset} & \textbf{Model} & \textbf{Resolution} & \textbf{Training Set} & \textbf{Test Set} & \textbf{Classes} & \textbf{Sample Size} \\
\midrule
\rowcolor{gray!10} MNIST & LeNet-5 & 28$\times$28 (Gray) & 60,000 & 10,000 & 10 & 1,000 (10\%) \\
CIFAR-10 & VGG16 & 32$\times$32 (RGB) & 50,000 & 10,000 & 10 & 1,000 (10\%) \\
\rowcolor{gray!10} ImageNet-1K & EfficientNetB3 & 300$\times$300 (RGB) & 1,200,000 & 50,000 & 1,000 & 1,000 (2\%) \\
\bottomrule
\end{tabular}
    \vspace{1mm}
\end{table*}

\subsection{DL Test Input Generators}
\label{subsec:TIG}
We selected the four TIGs outlined in Section \ref{sec:back_TIGs} because they encompass (1) diverse test generation approaches, i.e., PBA and GMA; (2) different access levels, i.e., white-box and black-box; (3) various test objectives, i.e., including misbehaviors, neuron coverage, and surprise coverage; and (4) they are open-source, allowing us to execute and evaluate them.

We applied modifications to the selected TIG when necessary to ensure a fair and comprehensive comparison. For the ImageNet1K \cite{krizhevsky2012imagenet} experiments, three TIGs required adjustments: For DeepHunter, we modified the profiler to handle higher-resolution images ($300 \times 300$ pixels) and EfficientNet-specific layers. We also updated the preprocessing steps to match EfficientNet’s input format and adjusted the layer selection to skip layers with less informative neuron coverage, allowing DeepHunter to effectively capture relevant activations in this model. DeepFault was adapted to handle higher-resolution images ($300 \times 300$ pixels) and to interface with EfficientNetB3, while AdvGAN’s generator network was scaled to handle the higher-resolution input of EfficientNetB3, ensuring compatibility without changing its core structure. Despite similar efforts to adapt SINVAD's Variational Autoencoder (VAE) architecture to accommodate the higher resolution required by EfficientNetB3, it proved insufficient to handle the complexity of ImageNet1K. SINVAD's VAE architecture could not be successfully adapted to process the higher-resolution ImageNet1K inputs. For the experiments involving MNIST and CIFAR-10 datasets, all tools retained their original implementations.

\subsection{Generation of Synthetic Test Inputs}

In this step, for each of the pre-trained DL models and the datasets described in Subsection \ref{sec:models_and_datasets}, we leverage the TIGs outlined in Subsection \ref{subsec:TIG} to generate synthetic test inputs. From each dataset's test set, we create 10 non-overlapping folds of 100 samples each, ensuring that all selected samples are correctly classified by the model under test. Each fold is treated as a separate experiment, maintaining independence between folds.
For each fold, we apply the TIGs to generate synthetic test cases. For each TIG, we construct a new synthetic collection of test cases, $D_{syn}$, where the number of generated images varies according to each TIG's generation strategy. $D_{syn}$ is then used to evaluate the performance of the TIGs. Some TIGs, such as DeepHunter and DeepFault, can be executed with different configurations, i.e., DeepHunter can operate using six different coverage criteria, while DeepFault can utilize three different suspiciousness measures (refer to Section \ref{subsec:TIG} for more details). We run these TIGs across all configurations and record the average values.
This experimental design aligns with the guidelines in ``Empirical Standards for Software Engineering Research"~\cite{ralph2020empirical}. In practice, when comparing approaches that incorporate random components, it is essential to perform multiple runs and apply statistical significance tests. The inherent randomness of these TIGs necessitates several runs to validate the tests' outcomes. For parameter tuning in each approach, we maintain the default values provided by the respective TIGs. We acknowledge that exploring the impact of different parameter settings in each approach represents an interesting direction for future work.

\subsection{Evaluation Criteria}

This section describes the metrics used to evaluate the generated inputs and assess the effectiveness of the TIGs. We employ five complementary metrics: Defect Detection Rate (DDR) \cite{eniser2019deepfault} \cite{wei2021deephunter}, Attack Success Rate (ASR) \cite{wu2021performance}, Learned Perceptual Image Patch Similarity (LPIPS) \cite{zhang2018unreasonable}, Perturbation Magnitude (PM) \cite{shaeiri2020towards}, and Execution Time (ET).

\textbf{Defect Detection Rate (DDR)}. The DDR quantifies a TIG's ability to expose robustness issues by calculating the proportion of original test inputs that, after transformation, lead to misclassification. DDR measures the TIG’s capability to make original inputs challenging for the model, providing insight into the overall robustness of the model when exposed to transformed examples. Formally, the DDR is defined as:

\begin{equation} 
\mathrm{DDR} = \frac{N_{\text{misclassified original}}}{N_{\text{total original}}} 
\end{equation}
where $N_{\text{misclassified original}}$ is the number of original test inputs that result in misclassification after transformation by the TIG, and $N_{\text{total original}}$ is the total number of original test inputs.

\textbf{Attack Success Rate (ASR)}. The ASR measures the success rate of the adversarial examples generated by the TIGs in causing misclassification. ASR focuses on the effectiveness of the adversarial examples produced, providing insight into how often the generated examples successfully cause misclassification. Formally, the ASR is defined as:

\begin{equation} 
\mathrm{ASR} = \frac{N_{\text{successful adversarial}}}{N_{\text{total generated}}} 
\end{equation}
where $N_{\text{successful adversarial}}$ is the number of adversarial examples that result in misclassification, and $N_{\text{total generated}}$ is the total number of generated test inputs. While both metrics assess a TIG's ability to induce misclassification, they focus on different aspects. DDR measures the proportion of original test inputs for which the TIG can find at least one misclassification. It emphasizes coverage across all original inputs, evaluating how effectively the TIG challenges each one. ASR, in contrast, calculates the overall success rate of all generated test inputs in causing misclassification, regardless of how many original inputs they originate from. By comparing DDR and ASR, we understand both the TIG's thoroughness in covering the test set (DDR) and the effectiveness of its generated inputs in inducing misclassification (ASR).

\textbf{Learned Perceptual Image Patch Similarity (LPIPS)}. To evaluate the naturalness and quality of the generated test cases, we employ LPIPS \cite{zhang2018unreasonable} metric. LPIPS calculates the perceptual distance between images based on deep features extracted from a neural network, providing a semantically meaningful assessment of image similarity. Specifically, we utilize the AlexNet architecture~\cite{krizhevsky2012imagenet} for feature extraction due to its effectiveness in capturing perceptual differences. This metric ensures that the generated test cases are perceptually distinct from the original inputs while avoiding nonsensical images that are not useful for robustness testing.


\textbf{Perturbation Magnitude (PM).} The PM measures the extent of changes applied to the original inputs to create synthetic inputs. It is calculated as the mean L2 norm of the difference between the original and transformed images, offering a quantitative assessment of the distortion introduced by the TIGs. Formally, PM is defined as: 
\begin{equation}
\mathrm{PM}=\frac{1}{N} \sum_{i=1}^N\left\|x_i^{\text {orig }}-x_i^{\mathrm{trans}}\right\|_2
\end{equation}
where $N$ is the total number of pixels in the image, $x_i^{\text{orig}}$ is the value of the $i$th pixel in the original image, and $x_i^{\mathrm{trans}}$ is the corresponding pixel value in the transformed image. \\
In our experiments, we compute (1) the average PM (Avg PM) and (2) the standard deviation PM (Std PM) across all images. The Avg PM indicates the overall level of change introduced by the TIG, while the Std PM reflects the diversity of the TIG transformations. 

\textbf{Execution Time (ET).}  To further evaluate the TIGs, we measured the execution time per image, calculated as the total processing time divided by the number of images tested. This metric captures the efficiency of each TIG, allowing us to assess their practicality in real-time or resource-limited scenarios.

By combining DDR, ASR, LPIPS, PM,  and ET, we provide a comprehensive evaluation of the TIGs. The DDR and ASR assess the TIGs' ability to reveal robustness issues and generate effective adversarial examples, respectively. The LPIPS assesses the perceptual similarity between the original and generated inputs, evaluating the quality of the generated test inputs. The PM quantifies the level of perturbation introduced to the original inputs, providing a quantitative assessment of the extent and diversity of transformations applied. ET captures the efficiency of each TIG.

\section{Results}
\label{sec:results}




In this section, we first briefly introduce the experimental environment, and then we detail the experiments and results obtained to answer our RQs.
All experiments were conducted on Compute Canada’s infrastructure, leveraging Intel E5-2683 v4 Broadwell processors and NVIDIA A100 GPUs running on a Linux operating system.

\begin{table*}[t]
    \centering
    \caption{Comparative Performance of TIGs Across Datasets, with bolded values indicating the best performance per metric within each dataset, highlighting each tool’s effectiveness across Detection Rate (DDR), Attack Success Rate (ASR), Perturbation Magnitude (PM), and Perceptual Similarity (LPIPS) in varying dataset complexities.}
    \renewcommand{\arraystretch}{1.2}  
    \begin{tabular}{lccccccc}
\toprule
\rowcolor{gray!10}
& & \multicolumn{2}{c}{\textbf{RQ1: Robustness}} & \textbf{RQ2: Naturalness} & \multicolumn{2}{c}{\textbf{RQ3: Diversity}} & \textbf{RQ4: Efficiency} \\
\rowcolor{gray!10}
\textbf{Tool} & \textbf{Dataset} & \textbf{DDR} & \textbf{ASR} & \textbf{LPIPS} & \textbf{PM (Avg)} & \textbf{PM (Std)} & \textbf{Time(s)} \\
\midrule
\multirow{3}{*}{Deephunter} 
    & MNIST      & 71.4\% & 38.24\% & 0.45 & 2.50 & 0.64 & 2.15 \\
    & CIFAR-10   & 72.04\% & 5.97\% & \textbf{0.23} & \textbf{4.47} & 1.81 & 46.34 \\
    & ImageNet-1k & 34.0\% & 2.27\% & \textbf{0.22} & 12.26 & \textbf{4.56} & 1,080.00 \\
\midrule
\multirow{3}{*}{Deepfault}
    & MNIST      & 1.0\% & 1.00\% & 0.17 & \textbf{0.13} & 0.017 & \textbf{0.178} \\
    & CIFAR-10   & \textbf{90.0\%} & \textbf{90.0\%} & 0.61 & 7.85 & 0.51 & \textbf{0.974} \\
    & ImageNet-1k & 42.6\% & 42.6\% & 0.44 & 5.79 & 0.3 & 743.35 \\
\midrule
\multirow{3}{*}{AdvGAN}
    & MNIST      & \textbf{99.1\%} & \textbf{99.1\%} & \textbf{0.12} & 3.83 & 0.52 & 15.60 \\
    & CIFAR-10   & 83.8\% & 83.8\% & 0.29 & 5.34 & 0.72 & 25.20 \\
    & ImageNet-1k & 32.0\% & 32.0\% & 0.30 & \textbf{1.28} & 0.1 & \textbf{60.00} \\
\midrule
\multirow{3}{*}{SinVAD}
    & MNIST      & 89.0\% & 89.0\% & 0.29 & 9.41 & \textbf{1.45} & 57.60 \\
    & CIFAR-10   & 52.3\% & 80.2\% & 0.56 & 28.75 & \textbf{4.93} & 87.00 \\
    & ImageNet-1k & \textbf{100.0\%} & \textbf{100.0\%} & 0.94 & \textit{inf} & 0.0 & 540.00 \\
\bottomrule

\multicolumn{8}{l}{\footnotesize{All execution times are reported in seconds per image.}}
\end{tabular}
    
    \label{tab:performance_comparison}
\end{table*}

\subsection{RQ1: Which TIG reveals more DNN robustness issues?}
\label{sec:Robustness}

\noindent
\textbf{Motivation.} The goal is to evaluate the effectiveness of state-of-the-art TIGs in revealing DNN robustness issues by identifying misclassified synthetic test cases, thus exposing potential vulnerabilities in the DL models.

\noindent{\textbf{Method.}} We generated synthetic test cases for each TIG from the original test set and computed the average Defect Detection Rate (DDR) and Attack Success Rate (ASR) across 10 runs. Since we are working with averaged metrics, to assess whether the differences across approaches are statistically significant, we use the Wilcoxon statistical test \cite{wilcoxon1992individual} and Vargha-Delaney  \cite{vargha2000critique} ($\hat{A}_{12}$) effect size test. The Wilcoxon statistical test determines if the difference between two means is statistically significant ($p$-value $< 0.05$). Vargha-Delaney $\hat{A}_{12}$ determines the magnitude of difference between two groups, with a range of [0, 1]. $\hat{A}_{12} > 0.5$ implies that values in the first group are larger, $\hat{A}_{12} < 0.5$ implies that they are lower, and $\hat{A}_{12} = 0.5$ indicates statistically indistinguishable groups. We used the coefficients proposed by Hess et al.~\cite{hess2004robust} to interpret the magnitude of the differences into negligible, small, medium, and large differences.

\textbf{Results:} 
Table \ref{tab:performance_comparison} summarizes DDR and ASR results across datasets. Results reveal clear differences in the robustness-revealing capabilities of TIGs across datasets and models.

For MNIST, GMAs (AdvGAN and SinVAD) significantly outperformed PBAs (DeepHunter and DeepFault). AdvGAN achieved the highest DDR/ASR (99.1\%), followed by SinVAD (89.0\%), due to their generative capabilities. In contrast, PBAs like DeepHunter had moderate performance (DDR: 71.4\%, ASR: 38.24\%), limited by their small-scale perturbations. DeepFault exhibited the weakest performance (DDR/ASR: 1.0\%), suggesting a mismatch between its fault localization strategy and the feature-simple dataset. 
For CIFAR-10, DeepFault excelled (DDR/ASR: 90.0\%), effectively using targeted perturbations in a feature-rich dataset. AdvGAN performed well (83.8\%), while SinVAD achieved a high ASR (80.2\%) but a lower DDR (52.3\%), indicating a focus on adversarial examples over defect detection. DeepHunter scored a low ASR (DDR: 72.04\%, ASR: 5.97\%), suggesting that its perturbation techniques may struggle with feature-rich datasets. 
For ImageNet-1k, performance declined across all TIGs. While SinVAD achieved a DDR/ASR of 100\%, it generated invalid test inputs (e.g., black images), highlighting an overfitting issue to simplistic patterns, which will be further discussed in RQ2. DeepFault showed consistent performance (DDR/ASR: 42.6\%), while AdvGAN and DeepHunter achieved similar DDRs (32–34\%), with DeepHunter’s ASR being particularly low (2.27\%), highlighting its ineffectiveness with high-resolution data.

Statistical analysis confirmed significant differences among TIGs (p-value $< 0.05$), with a large effect size ($\hat{A}_{12} > 0.5$) in most cases.

\begin{tcolorbox}[colback=blue!5,colframe=blue!40!black]

\textbf{Findings:} GMA TIGs (AdvGAN and SinVAD) excel on simpler datasets but face challenges with complex datasets. DeepFault performs consistently, especially on complex datasets. DeepHunter, constrained by its small perturbations, is less effective as dataset complexity increases.

\noindent{\textbf{Challenges:}} SinVAD generated Out-of-Distribution inputs for ImageNet-1k, struggling to produce valid examples.

\end{tcolorbox}

\subsection{RQ2: Which TIG generates more natural test cases?}
\label{sec:naturalness}

\noindent
\textbf{Motivation.} The purpose of test cases is to identify DL erroneous behavior that could potentially occur in practice. Hence, assessing the naturalness of synthetically generated test cases is fundamental to infer their usefulness in real-world scenarios.

\noindent{\textbf{Method.}} We evaluate the naturalness of test cases using the Learned Perceptual Image Patch Similarity (LPIPS). LPIPS provides a perceptual similarity score between original and generated images, where lower values indicate more natural-looking images.
 
\textbf{Results:} 

The fifth column of Table \ref{tab:performance_comparison} reports LPIPS findings, revealing that TIG naturalness is highly dataset-dependent and influenced by model architecture. 

For MNIST, AdvGAN achieved the lowest LPIPS score (0.12), indicating that it generated the most natural-looking test cases. This aligns with GAN’s ability to learn the underlying data distribution and produce realistic perturbations that are perceptually similar to the original inputs. DeepFault and SinVAD followed with LPIPS scores of 0.17 and 0.29, respectively, showing relatively natural perturbations. In contrast, DeepHunter had the highest LPIPS score (0.45), suggesting less natural test cases, likely due to more visible, pixel-level changes. For CIFAR-10, DeepHunter produced the most natural test cases (LPIPS: 0.23), indicating that its perturbation strategy aligned well with the dataset’s features. AdvGAN performed comparably (0.29), suggesting only minor perceptual differences. However, SinVAD and DeepFault had higher LPIPS scores (0.56 and 0.61), indicating a drop in naturalness. This decline can be attributed to the internal functioning of these TIGs: SinVAD relies on a VAE, which struggled with the rich features of CIFAR-10, leading to less realistic outputs. DeepFault, with its targeted neuron activation strategy, applied more aggressive modifications that disrupted the natural appearance of the test cases. For ImageNet-1k, DeepHunter again led in naturalness (LPIPS: 0.22), followed by AdvGAN (0.30). DeepFault showed a degradation in naturalness (0.44), while SinVAD had the highest LPIPS score (0.94), indicating the least natural results. The poor performance of SinVAD on this dataset can be attributed to the VAE’s difficulty in generating meaningful inputs, as the generative model struggled to learn the complex features of ImageNet-1k, resulting in unrealistic outputs.

Overall, an inverse correlation between DDR/ASR and naturalness scores was observed. High DDR and ASR often correspond with lower naturalness, as more aggressive modifications are typically required to induce misclassifications, particularly in complex datasets.

\begin{tcolorbox}[colback=blue!5,colframe=blue!40!black]
\textbf{Findings:} GMAs (AdvGAN and SinVAD) tend to produce more natural modifications for simpler datasets, while PBAs like DeepHunter excel in generating natural-looking test cases for more complex datasets. An inverse relationship between robustness-revealing metrics (DDR/ASR) and naturalness was observed.

\end{tcolorbox}

\subsection{RQ3: Which TIG generates more diversified test cases?}
\textbf{Motivation.} Diversified test cases are crucial to evaluate DL models in broader input space and under different situations. This RQ aims to assess each TIG's ability to generate varied test cases that explore different parts of the input space.

\textbf{Method:} We evaluate the diversity of the generated test cases using the Perturbation Magnitude (PM) metric, which quantifies the extent of modifications made to induce model misbehavior. Two key measures are considered: PM Avg, representing the average magnitude of perturbations applied to the original images, where higher values indicate stronger perturbations, and PM Standard deviation (Std), reflecting the variability in perturbation magnitude across the generated inputs, with higher values indicating greater diversity in the test cases.

\textbf{Results:} 

The sixth and seventh columns of Table \ref{tab:performance_comparison} report Perturbation magnitude average (PM Avg) and Perturbation magnitude standard deviation (PM Std) findings, highlighting differences in perturbation strength and diversity across TIGs.

For MNIST, SinVAD had the highest PM Avg (9.41) and PM Std (1.45), reflecting strong and varied perturbations, making it the most diverse TIG. DeepHunter followed with a moderate PM Avg (2.50) and a high PM Std (0.64), indicating diverse but controlled perturbations. AdvGAN showed less diversity (PM Std: 0.52) but higher perturbation strength (PM Avg: 3.83). DeepFault exhibited minimal and uniform changes, with the lowest PM Avg (0.13) and PM Std (0.17). For CIFAR-10, SinVAD again led in perturbation strength (PM Avg: 28.75) and diversity (PM Std: 4.93). DeepHunter achieved a balance with a moderate PM Avg (4.47) and higher PM Std (1.81). In contrast, AdvGAN (PM Avg: 5.34, PM Std: 0.72), and DeepFault (PM Avg: 7.85, PM Std: 0.51) applied stronger but less varied perturbations. For ImageNet-1k, SinVAD failed, producing extreme and uniform perturbations (PM Avg: $\infty$, PM Std: 0.0), generating out-of-distribution data. DeepHunter achieved a balanced performance (PM Avg: 12.26, PM Std: 4.56), with moderate perturbation strength and diversity. AdvGAN (PM Avg: 1.28, PM Std: 0.1) and DeepFault (PM Avg: 5.79, PM Std: 0.3) exhibited limited diversity and weaker perturbations, struggling with the complexity of ImageNet-1k.

\begin{tcolorbox}[colback=blue!5,colframe=blue!40!black]
\textbf{Findings:} SinVAD shows the highest diversity across simpler datasets but fails on ImageNet-1k. DeepHunter consistently balances perturbation strength and diversity across all datasets, making it the most reliable TIG. AdvGAN and DeepFault apply stronger but less diverse perturbations, indicating a focus on generating aggressive perturbations rather than exploring varied test cases.  
\end{tcolorbox}

\subsection{RQ4: Which TIG is more efficient in test case generation?}

\noindent{\textbf{Motivation.}} A TIG's efficiency is essential for promoting its adoption in real-world software testing scenarios. Efficient TIGs enable faster defect detection and optimize resource usage. This RQ evaluates the performance of each TIG while generating test inputs.

\noindent{\textbf{Method.}} We measured the execution time of each TIG across three datasets (MNIST, CIFAR-10, and ImageNet-1k). Each tool was run on 10 seed inputs, and the average execution time was calculated. To ensure comparability, execution times were normalized to a per-image basis, and all times were standardized to seconds.

\noindent{\textbf{Results:}} The eighth column of Table \ref{tab:performance_comparison} reports the normalized execution times (in seconds per image) for each TIG across the three datasets.

DeepFault achieved the fastest execution on MNIST (0.178 s/image) and CIFAR-10 (0.974 s/image). However, its performance degraded significantly on ImageNet-1k (743.35 s/image), due to (1) the increased image resolution in ImageNet-1k ($300 \times 300$) and (2) the computational demands when analyzing the more complex EfficientNet-B3 model. AdvGAN showed the highest scalability across all datasets, with relatively low execution times: 15.60 s/image for MNIST, 25.20 s/image for CIFAR-10, and only 60 s/image for ImageNet-1k. This consistent performance can be attributed to its generative model, which produces adversarial examples in a single forward pass without iterative optimization, making it highly efficient even on large datasets. In contrast, SinVAD exhibited the highest execution times on smaller datasets (57.60 s/image for MNIST and 87.00 s/image for CIFAR-10), primarily due to the overhead of training its VAE model. Despite this, SinVAD's execution time became more competitive for ImageNet-1k (540 s/image), reducing the relative impact of its initial overhead. DeepHunter showed competitive performance on smaller datasets, with execution times of 2.15 s/image for MNIST and 46.34 s/image for CIFAR-10. However, it exhibited a drastic increase to 1080 s/image on ImageNet-1k. This substantial increase can be explained by its reliance on coverage-guided fuzzing, which involves extensive iterative modifications and evaluation steps.

\begin{tcolorbox}[colback=blue!5,colframe=blue!40!black]
\textbf{Findings:} AdvGAN was the most efficient overall, with consistent performance across all datasets. DeepFault excelled on smaller datasets but faced scalability issues on ImageNet-1k. SinVAD’s overhead affected its performance on small datasets but its overall performance improved on larger data. DeepHunter was efficient on simpler datasets but had significant delays on ImageNet-1k due to its iterative fuzzing approach.
\end{tcolorbox}

\begin{figure*}[t!]  
    \centering
    \begin{subfigure}[b]{0.32\textwidth}
        \centering
        \includegraphics[width=\textwidth]{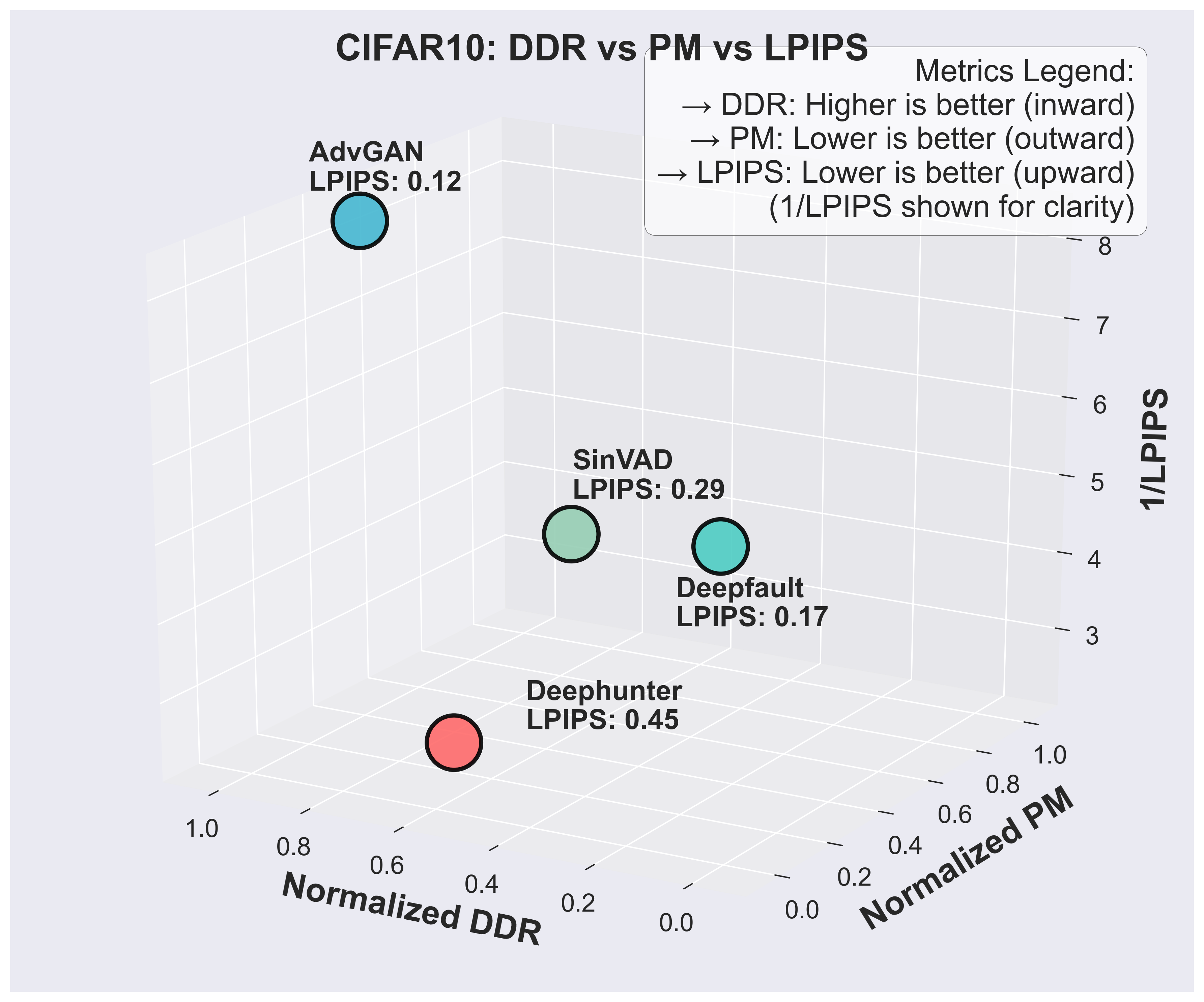}
        \caption{3D Performance on MNIST}
        \label{fig:mnist}
    \end{subfigure}\hfill
    \begin{subfigure}[b]{0.32\textwidth}
        \centering
        \includegraphics[width=\textwidth]{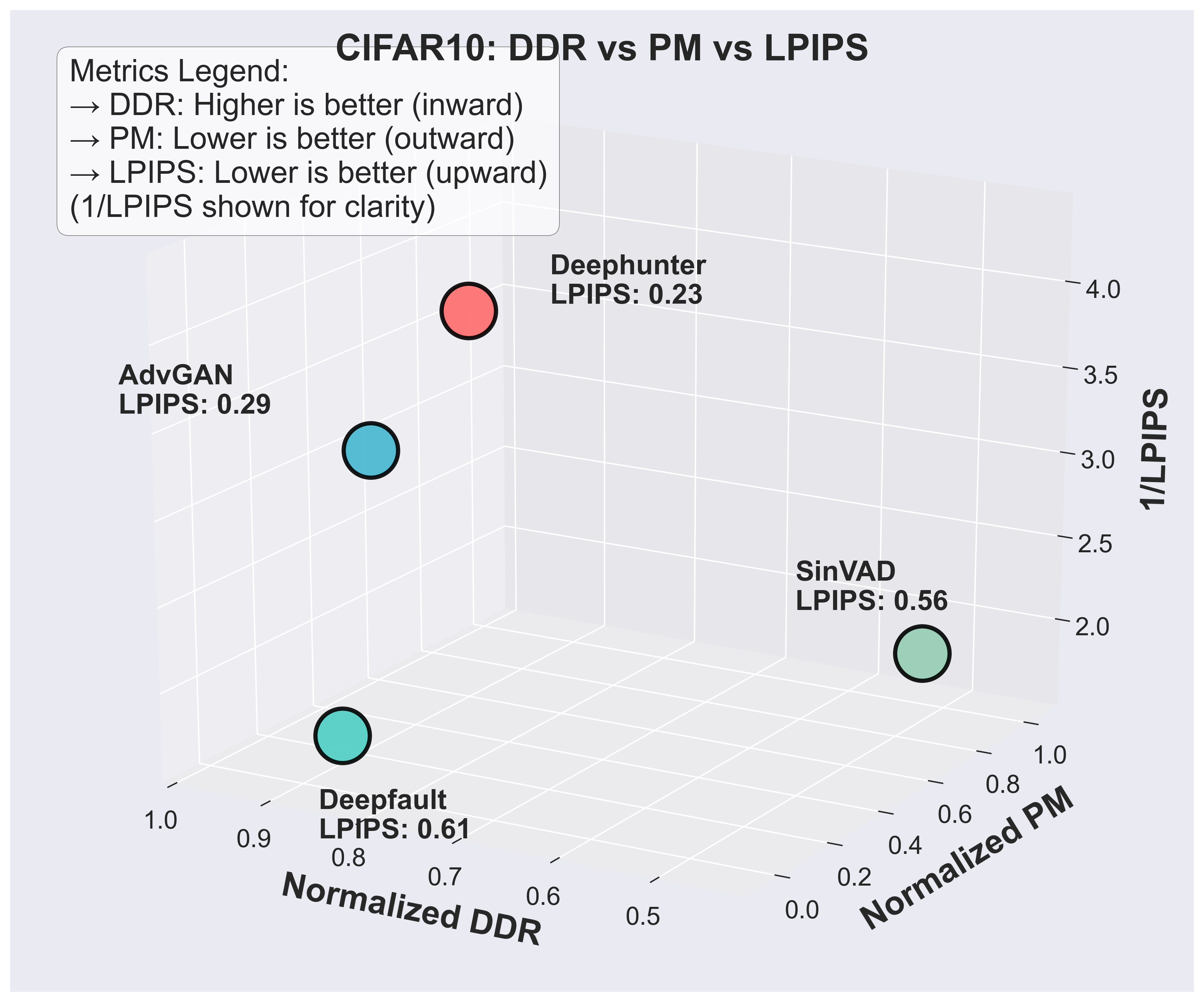}
        \caption{3D Performance on CIFAR-10}
        \label{fig:cifar}
    \end{subfigure}\hfill
    \begin{subfigure}[b]{0.32\textwidth}
        \centering
        \includegraphics[width=\textwidth]{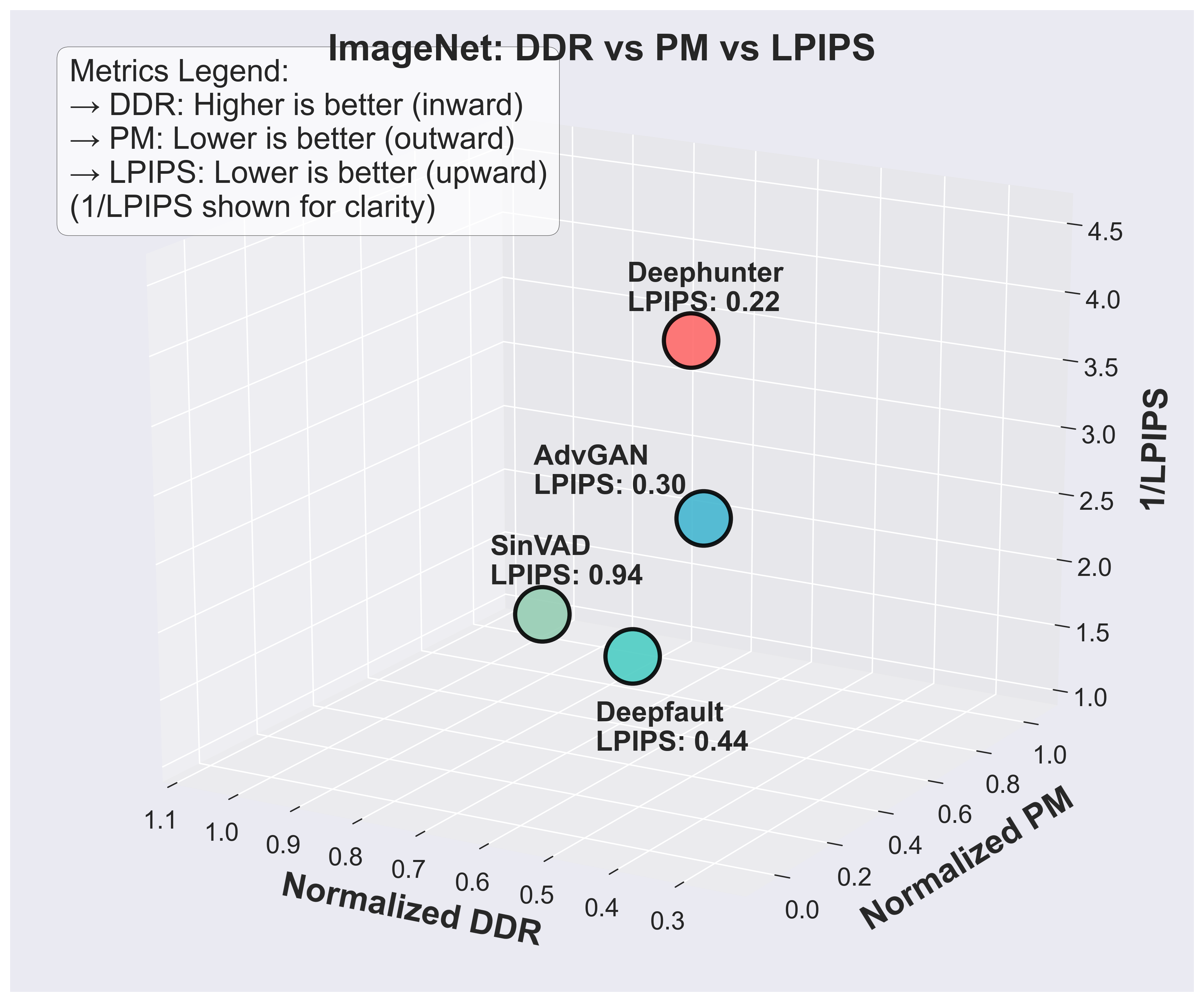}
        \caption{3D Performance on ImageNet}
        \label{fig:imagenet}
    \end{subfigure}
    \caption{\small 3D Performance Comparison of DNN Testing Tools, illustrating Detection Rate (DDR), Perturbation Magnitude (PM), and Perceptual Similarity (LPIPS) metrics across increasing dataset complexity in ImageNet, CIFAR-10, and MNIST, highlighting each tool's effectiveness and robustness.}
    \label{fig:performance_comparison}
\end{figure*}

\section{Discussion}
\label{sec:discussion}
\subsection*{Assessing TIGs: Strengths, Weaknesses, and Paths Forward}
The evaluation of TIGs in our study highlights a range of distinct strengths, limitations, and inherent trade-offs among the different approaches \ref{fig:performance_comparison}. AdvGAN demonstrates a strong balance between revealing robustness issues and maintaining natural perturbations on simpler datasets. However, it struggles to scale effectively with high-resolution and complex data. Its strength lies in generating minimal perturbations, though this comes at the cost of reduced effectiveness on more complex datasets. Enhancing the generator and discriminator architectures could be a viable path to improving its efficacy on complex data. SinVAD, known for its ability to generate highly diverse test cases, also faces limitations when applied to high-resolution images due to the inherent limitations of its VAE architecture. The standard VAE employed in SinVAD lacks the expressive capacity to effectively model the complex data distributions found in high-resolution datasets. As a result, the generated samples often fall outside the valid data distribution, leading to unrealistic or out-of-distribution (OOD) test cases. To mitigate this issue, incorporating advanced generative models like Vector Quantized Variational Autoencoder (VQ-VAE) \cite{van2017neural} could enhance the latent space representation. VQ-VAE, with its discrete latent codebook and improved expressiveness, may enable SinVAD to better capture the intricate data patterns of high-resolution images, thereby producing more realistic and in-distribution test cases.

DeepFault and DeepHunter offer complementary strengths: DeepHunter excels in generating natural test cases, whereas DeepFault is more effective in exposing robustness issues. These tools show variable effectiveness across dataset complexities, underscoring a trade-off between robustness-revealing capability and maintaining natural perturbations. To address this trade-off, future work could focus on enhancing adaptive perturbation strategies, as well as leveraging recent advances in generative modeling.

\subsection*{Adaptation Challenges and Practical Considerations for TIGs}

The process of adapting TIGs to different models and datasets revealed several practical challenges, particularly with newer, more complex architectures like EfficientNetB3. Many TIGs, including those evaluated in this study, were initially designed for simpler, fully connected models and older datasets. This legacy focus requires significant adaptation to accommodate modern architectures. Specifically, TIGs like DeepHunter rely on profiling methods tailored to earlier, shallower networks, which failed to work effectively with more recent models. Adjustments were required to enable these profiling techniques to function properly with current architectures.

\begin{figure}[ht]
    \centering
    \includegraphics[width=1.0\columnwidth]{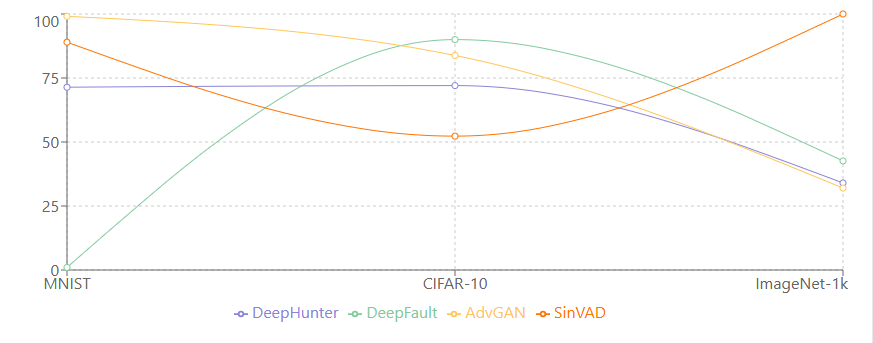}
    \caption{DDR Performance Trends of TIGs with Increasing Dataset Complexity, illustrating the scalability challenges encountered on more complex datasets.}
    \label{fig:trendplot}
\end{figure}

Furthermore, the lack of modular design in many TIG codebases complicated the process of adapting them to new datasets. When switching from the default dataset, numerous changes were needed across various code files, making the adaptation process laborious. This experience highlights the importance of a modular architecture in TIGs, allowing users to easily reconfigure the system for different datasets and models without extensive code alterations. Finally, as illustrated in Figure~\ref{fig:trendplot}, the performance of TIGs tends to decline as the complexity of the dataset increases. This observation suggests an urgent need for the development of more robust and adaptable TIGs that can effectively handle the demands of modern, complex model-dataset pairs.

\subsection*{More naturalness metrics are required}
The relevance of a test case is directly proportional to the likelihood that the scenario it captures will occur in a real-world setting. Since DL systems are designed to handle real-world events, a test that induces erroneous behavior is only useful if it remains within realistic bounds; otherwise, its practical value is limited. Therefore, metrics and techniques for assessing and preserving the naturalness of generated test cases are crucial parts of every test generation approach's workflow. In this work, we used LPIPS as a proxy for naturalness. However, other naturalness measurements exist in the literature such as Inception Score~\cite{salimans2016improved} and Image Quality Assessment metrics~\cite{wang2009mean, wang2002universal} and might provide contradictory findings since they portray naturalness from distinct perspectives. Measuring the visual quality of an image is very subjective, and there is no precise solution yet for formalizing its assessment, making naturalness a hot topic that requires further exploration. To address the limitations of automated metrics, researchers have used \cite{hu2022if} human evaluations rather than these metrics to quantify naturalness. However, because using humans is not always feasible (e.g., in large-scale tasks), an automated approach must be developed.

\subsection*{Leveraging Large Language Models (LLMs) for Enhanced  TIGs}

LLMs offer promising opportunities for automating the generation of complex, diverse test inputs. Their generative power can efficiently produce edge cases and domain-specific inputs, reducing manual efforts and enhancing the realism of test cases, which may help identify vulnerabilities more effectively. Recent work, such as LANCE \cite{prabhu2023lancestresstestingvisualmodels}, has demonstrated the feasibility of using LLMs for generating test cases in image-based tasks. By leveraging large language modeling and text-based image editing, LANCE expands the range of test scenarios and highlights the potential of LLMs to enhance TIGs.

Looking ahead, LLMs could be utilized to create initial batches of diverse test cases, which traditional TIGs can then refine. This complementary strategy would leverage the strengths of both methods, increasing the diversity and complexity of test cases while minimizing computational overhead. Future work should explore deeper integration of LLMs to make TIGs more scalable and versatile, addressing a wider scope of robustness testing challenges.

\section{Related Work}
\label{sec:related-work}
As deep learning (DL) systems are used in critical applications \cite{inbook}, ensuring their robustness has driven advances in DL testing \cite{wang2024surveyneuralnetworkrobustness,huang2023hierarchicaldistributionawaretestingdeep,wang2021robotrobustnessorientedtestingdeep} approaches. Foundational work began with DeepXplore \cite{pei2017deepxplore}, using neuron coverage to detect model errors, and DeepTest \cite{tian2018deeptestautomatedtestingdeepneuralnetworkdriven}, which integrated metamorphic testing \cite{khokhar2020metamorphic} to evaluate model behavior in autonomous driving. TensorFuzz \cite{odena2018tensorfuzzdebuggingneuralnetworks} and DeepHunter \cite{xie2019deephunter} expanded coverage-guided testing by using fuzzing \cite{guo2018dlfuzz} and metamorphic mutation \cite{xie2019deephunter} to generate diverse input scenarios. Later, tools like DeepFault \cite{eniser2019deepfault} focused on fault localization, while DeepEvolution \cite{braiek2019deepevolution}, DiverGet \cite{yahmed2022diverget}, and DeepJanus \cite{riccio2020modelbasedexplorationfrontierbehaviours} introduced search-based approaches and boundary testing to explore decision thresholds. 
Coverage-guided methods \cite{xie2019deephunter, BOUCHOUCHA2023107281} have evolved to include neuron, layer, and surprise metrics \cite{kim2019guiding}, adapting testing for specific architectures like CNNs and Transformers \cite{islam2024comprehensive}. Adversarial testing \cite{huang2020survey} has progressed from generating adversarial examples \cite{goodfellow2014explaining} to black-box testing, with frameworks integrating robustness evaluation and adversarial defenses \cite{ren2020adversarial}.
Generative models now play a role, with GANs \cite{goodfellow2020generative} and VAEs \cite{pinheiro2021variational} creating realistic test cases. DeepRoad \cite{zhang2018deeproadganbasedmetamorphicautonomous} used GANs \cite{goodfellow2020generative} for autonomous driving tests, while recent work with StyleGAN \cite{karras2019stylebasedgeneratorarchitecturegenerative}, diffusion models \cite{rombach2022highresolutionimagesynthesislatent} and adversarial generators such as AdvGAN \cite{xiao2018generating} and SinVAD \cite{kang2020sinvad} provides high-fidelity test inputs across domains. 

However, a clear gap remains in the literature regarding a comprehensive assessment of current TIGs across multiple performance dimensions. This gap leaves practitioners without practical guidelines for selecting or leveraging TIGs to their specific needs.

\section{Threats to Validity}
\label{sec:disc-thre-valid}

In the following, we discuss the threats to the validity of our study.

\textbf{Threats to internal validity.} They may result from how the empirical study was conducted. To mitigate these issues, we used the default configurations of evaluated TIGs described in the original paper as the performance of TIGs can vary depending on the parameters selected. Additionally, to mitigate the risk of random variation affecting results, we repeated experiments 10 times. To confirm the statistical significance of our findings, we performed statistical hypothesis testing and effect size assessments using the non-parametric Wilcoxon test \cite{wilcoxon1992individual} and the Vargha–Delaney effect size \cite{vargha2000critique}.

\textbf{Threats to external validity.} They concern the applicability of our findings across different models, datasets, and TIGs. To mitigate these issues, we selected four state-of-the-art TIGs, three DL models of varying complexities, and three datasets of different sizes. This variety allowed us to evaluate the performance of TIGs in a broad spectrum of scenarios, from simple image datasets like MNIST to complex high-resolution datasets such as ImageNet-1K. However, our findings may not fully extend beyond image classification. Future work should assess TIGs in other domains, such as NLP or speech recognition, to enhance external validity.

\textbf{Threats to conclusion validity.} They involve the risk of bias arising from dataset and model selection and analysis procedures, potentially influencing the validity of the conclusions. To reduce selection bias, we chose well-known datasets (MNIST, CIFAR-10, ImageNet-1K) and popular DL architectures (LeNet-5, VGG16, EfficientNetB3) commonly used in the DL and software testing communities.

\section{Conclusion}
\label{sec:conclusion}

Our empirical study of four leading TIGs across datasets of varying complexity reveals significant performance variations across key metrics, including robustness detection (DDR, ASR), naturalness (LPIPS), diversity (PM), and computational efficiency. For example, while AdvGAN performed exceptionally well on MNIST (DDR/ASR: 99.1\%, LPIPS: 0.12), its effectiveness dropped sharply on ImageNet-1k (DDR/ASR: 32.0\%, LPIPS: 0.30). Similarly, DeepFault exhibited stark contrasts between MNIST (DDR/ASR: 1.0\%) and CIFAR-10 (DDR/ASR: 90.0\%). Increasing dataset complexity also posed challenges, leading to longer execution times and fluctuating effectiveness. DeepHunter’s runtime surged from 2.15s on MNIST to 1,080s per image on ImageNet-1k, while SinVAD’s naturalness deteriorated (LPIPS from 0.29 to 0.94) and failed to generate valid inputs for ImageNet-1k. These findings emphasize the importance of selecting TIGs based on dataset complexity, testing objectives, and resource constraints, highlighting the need for adaptable, modular testing frameworks. Future research should focus on developing scalable methodologies that address the rising complexity of DL systems and refining naturalness metrics to better capture perceptually meaningful variations.

\balance
\bibliographystyle{IEEEtran}
\bibliography{project.bib}
\end{document}